\documentclass[letterpaper]{article} 
\usepackage{aaai23}  
\usepackage{times}  
\usepackage{helvet}  
\usepackage{courier}  
\usepackage[hyphens]{url}  
\usepackage{graphicx} 
\usepackage{natbib}  
\usepackage{caption} 
\usepackage{algorithm}
\usepackage{algorithmic}

\usepackage{xcolor,soul}
\usepackage{mdframed}
\usepackage{tabularx}
\usepackage{tikz}
\usepackage{multirow}
\usepackage{newfloat}
\usepackage{listings}
\DeclareCaptionStyle{ruled}{labelfont=normalfont,labelsep=colon,strut=off} 
\floatstyle{ruled}
\newfloat{listing}{tb}{lst}{}
\floatname{listing}{Listing}
\title{Ground Manipulator Primitive Tasks to Executable
Actions \\using Large Language Models}
\author{
    Yue Cao and C.S. George Lee
}
\affiliations{
    Elmore Family School of Electrical
and Computer Engineering, Purdue University
\\
    \{yuecao, csglee\}@purdue.edu}
\usepackage{bibentry}

\usepackage[symbol]{footmisc}

\begin{document}

\maketitle

\begin{abstract}
Layered architectures have been widely used in robot systems. The majority of them implement planning and execution functions in separate layers. However, there still lacks a straightforward way to transit high-level tasks in the planning layer to the low-level motor commands in the execution layer. In order to tackle this challenge, we  propose a novel approach to ground the manipulator primitive tasks to robot low-level actions using large language models (LLMs). We designed a program-function-like prompt based on the task frame formalism. In this way, we enable LLMs to generate position/force set-points for hybrid control. Evaluations over several state-of-the-art LLMs are provided.
\end{abstract}
\footnotetext[2]{This work was supported in part by the National
Science Foundation under Grant IIS-1813935.
Any opinion, findings,
and conclusions or recommendations expressed in this material are
those of the authors and do not necessarily reflect the views of
the National Science Foundation.}
\section{Introduction}
In robotics, the layered architectures break down the system design into multiple layers, then separately configure each layer. The development of layered architectures can be traced back to the Stanford Research Institute's \textit{Shakey} mobile robot~\cite{Nilsson69}. Its system was programmed by high-level functions solving planning problems in first-order logic and low-level functions specifying motor commands. By the early 1990s, a number of layered architectures have been developed and summarized as the sense-model-plan-act (SMPA) paradigm in~\cite{Brooks91}. Nowadays, the layered architectures typically consisting of a planning layer and an execution player have been widely used in many robot systems~\cite{KortenkampSB16}. 

However, there is a long-lasting challenge in the layered system -- the transition from planning to execution. Back in 1987, Rodney Brooks had pointed out the drawbacks of separating planning and execution layers~\cite{brooks1987planning}. To deal with this issue, Brooks proposed a biological-system-inspired subsumption architecture~\cite{brooks1991new}. But after years, the subsumption architecture didn't achieve much progress in modern robot control. Most robot systems stay with the layered architectures and one common solution for the planning-to-execution transition problem is using primitive tasks. As early as in the \textit{Shakey} robot, it introduced a library of intermediate-level actions such as ``push object'' or ``go  through door'' to communicate between the high-level logic and low-level motor commands~\cite{Nilsson84}. A classic 3-layer architecture in~\cite{firby1989adaptive} was separated into high-level planning, middle-level execution, and low-level hardware control layers. Its middle-level layer implements a reactive-action-package (RAP) system~\cite{firby1992building} consisting of primitive tasks to bridge between the symbolic plans and low-level control actions. Nevertheless, the existing primitive-task-based approaches require laborious manual specification. They usually yield in some libraries of a small number of primitive tasks, thereby limiting the task adaptability and generalizability of robots. 
 
To seek a new method for the planning-to-execution transition problem, we turn our attention to the state of the art -- large language models (LLMs). The large language models are neural-network-based language
models with a huge number of parameters and trained over massive amount of data. They significantly outperform other natural language processing (NLP) models and exhibit remarkable capabilities in generating textual responses to a broad range of questions. Since the release of OpenAI ChatGPT, the emergent abilities of LLMs have drawn the interests of
researchers from a variety of disciplines, including robotics. In robotics, pilot studies have primarily focused on the planning layer~\cite{pmlr-v162-huang22a, ahn2022can, DBLP:conf/aaaiss/0007L23, xiang2023language, lin2023text2motion}. They utilized the large language models to generate high-level task plans in a zero shot or few shot manner. However, the generated abstract plans are described in natural languages, making them uninterpretable by the execution layer. 

\begin{figure}[htb]
  \centering
  \begin{tabular}{c@{\hspace{5pt}} c }
    \small (a) & \begin{minipage}{0.85\linewidth}\includegraphics[width=\linewidth]{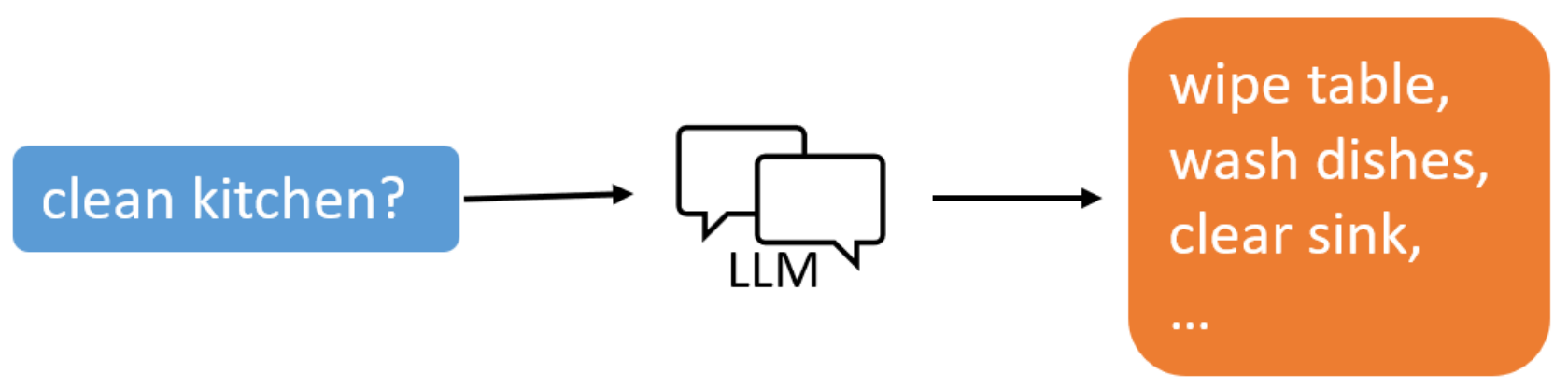}\end{minipage} \\
    \small (b) &\begin{minipage}{0.85\linewidth}\includegraphics[width=\linewidth]{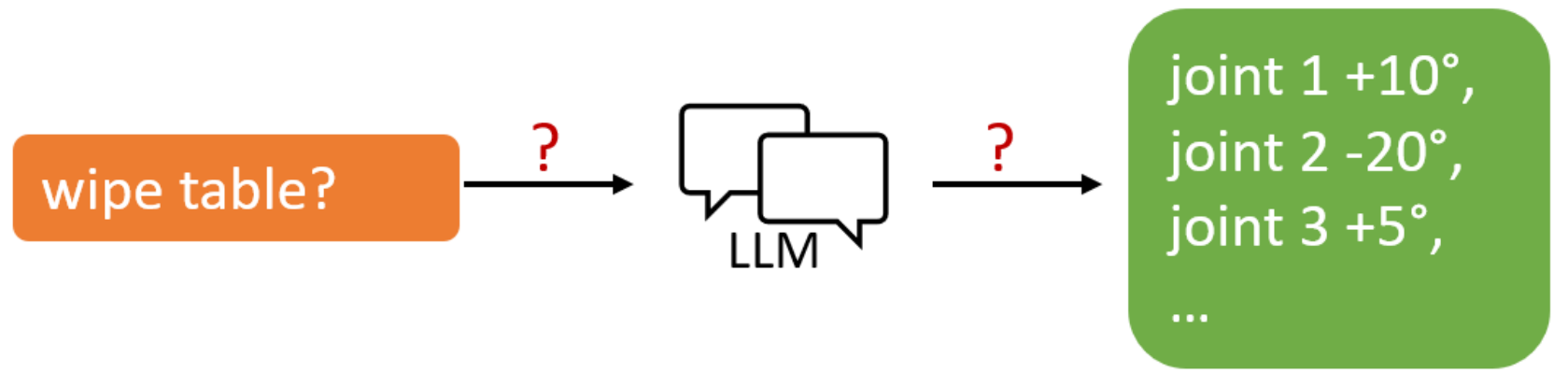}\end{minipage} 
  \end{tabular}

  \caption{(a) Previous work in LLM \& robotics mainly studied task generation in the planning layer. (b) Our proposed approach targets the planning-to-execution transition problem.}
\label{fig:LLMflow}
\vspace{-3mm}
\end{figure}

In this paper, we focus on the manipulator tasks and study the ability of LLMs to solve the planning-to-execution transition problem. Specifically, given a manipulator primitive task described in the natural language domain, we seek a LLM-based solution to translate it to low-level motor commands as shown in Figure~\ref{fig:LLMflow}.
We propose a novel approach to ground textual manipulator primitive tasks to executable actions using LLMs. Specifically, we design a program-like prompt based on the task frame formalism (TFF), an object-centric specification that facilitates task transfer across different manipulators. The prompt takes the text of a primitive task as input and outputs a set of position/force set-points in the task frame. These position/force set-points will allow manipulators to compute their low-level motor commands.

Major contributions of this paper are:
\begin{enumerate}
\item[(1)] We propose a LLM-based approach that enables language-described manipulator primitive tasks to be converted to the set-points for robot position/force hybrid control. It offers a new solution for the planning-to-execution transition problem in layered architectures of manipulators. 

\item[(2)] The remarkable generalizability of LLMs makes it possible to apply many manipulator primitive tasks described in natural language within our approach. Hence, it has the potential to reduce the burden on end-users when it comes to designing a complex library of primitive tasks.

\end{enumerate}

\section{Related Work}
\subsection{Language Grounding to Robot Actions}
Language grounding was introduced to robotics in~\cite{roy2005semiotic} to establish connections between natural languages and robot capabilities, including perception and action. Language grounding to action is generally referred to as associating high-level language commands with the low-level robot control system~\cite{DBLP:journals/aim/ChaiFLS16}. In the area of language grounding to action, the study of grounding schema follows one step behind the development of NLP techniques. 

The early work in~\cite{Kress-GazitFP08} enforced strict grammar rules to transform task specifications to linear temporal logic form. Later on, grammar parsers~\cite{TellexKDWBTR11, DBLP:conf/iser/MatuszekHZF12} such as the Stanford parser~\cite{MarneffeMM06} were applied to decompose sentence instructions into different grammatical elements, mainly involving verbs and objects. Subsequently, word-based statistical machine translation models were investigated in~\cite{squire2015grounding} to enable generalization to new environments. In the beginning of the deep-learning era, word embeddings and recurrent neural networks (RNNs) became the main steam and were later introduced in the robot grounding schema~\cite{ArumugamKGWRWT19, toyoda2021embodying}. 

Since the 2020s, large language models have achieved tremendous progress and become the primary focus of NLP. Using LLMs, one can bypass analyzing grammatical structure or word semantics, as was used in previous work.

\subsection{Task Frame Formalism}
The concept of the task frame formalism originated from the study of compliant
motions~\cite{mason1981compliance}. It was then explicitly formulated as an intermediate transfer between task planning and force/position control~\cite{bruyninckx1996specification}. The task frame, also known as the compliance frame, is a local coordinate frame attached to the object being manipulated. Its translational and rotational directions are configured to be either force controlled or position controlled. Such formalism allows end-users to implement hybrid position/force control strategy~\cite{raibert1981hybrid} alongside separate frame directions. 

Based on the task frame formalism, a variety of manipulator primitive tasks have been explored~\cite{BallardH86, morrow1997manipulation, kroger2004task, vuong2021learning}. Nevertheless, they mainly focused on the taxonomy of low-level action representation and provided little consideration to the association with high-level task nomenclature. For example, manipulator primitive tasks were specified as ``rotate about $y$ 5 deg'' or  ``translate along $x$ until next contact'' in~\cite{vuong2021learning}. But there is no straightforward way to translate natural-language-described tasks like ``insert key'' or ``open bottle'' to such low-level movements. 
\section{Proposed Approach}
\subsection{Preprocess: Identify Manipulator Primitive Tasks}
To start with, we need to ensure that the given primitive task conforms to the task frame formalism. For instance, we generally consider tasks like ``insert peg'' or ``open bottle''  as primitive tasks because they can be accomplished by a single control strategy without changing the coordinate setting. Tasks such as ``assembly GPU'' or ``make coffee'' are examples of non-primitive tasks because they must be further decomposed into multiple steps that use different control strategies and coordinates.

We use LLMs to assess the primitiveness of tasks. In fact, current LLMs including OpenAI ChatGPT, Google Bard, and Meta LLaMA-2, have some knowledge of the task frame formalism. For example, they can respond to the following prompt with reasonable answers.
\begin{mdframed}[backgroundcolor=gray!15]
Do you understand task frame formalism in robotics? 
\end{mdframed}

But when it comes to identifying the manipulator primitive tasks, their criteria become inconsistent. Therefore, we write a prompt including detailed explanation to query the LLM, as shown below.
\begin{mdframed}[backgroundcolor=gray!15]
We specify primitive tasks based on the concept of task frame formalism in robotics. \\
For example, ``open door'', ``slice bread'', ``insert card'' are primitive tasks because they use  a single control strategy and coordinate setting. While ``assembly computer'', ``make coffee'', ``drive car'' are not primitive tasks because they must be decomposed into several low-level tasks in  single control strategies and coordinate settings. \\
So, is ``open bottle'' a primitive task? Just answer yes or no.
\end{mdframed}

With such prompt, the LLMs can effectively determine whether a given manipulator task is primitive or not. If a certain task is not primitive, we need to perform task decomposition. Since the task decomposition is out of the scope of this
work, we refer to other literature~\cite{pmlr-v162-huang22a, ahn2022can, DBLP:conf/aaaiss/0007L23}.  

\subsection{TFF-based Prompt Design}
\begin{figure}[hbt]
     \centering
     \includegraphics[width=0.95\linewidth]{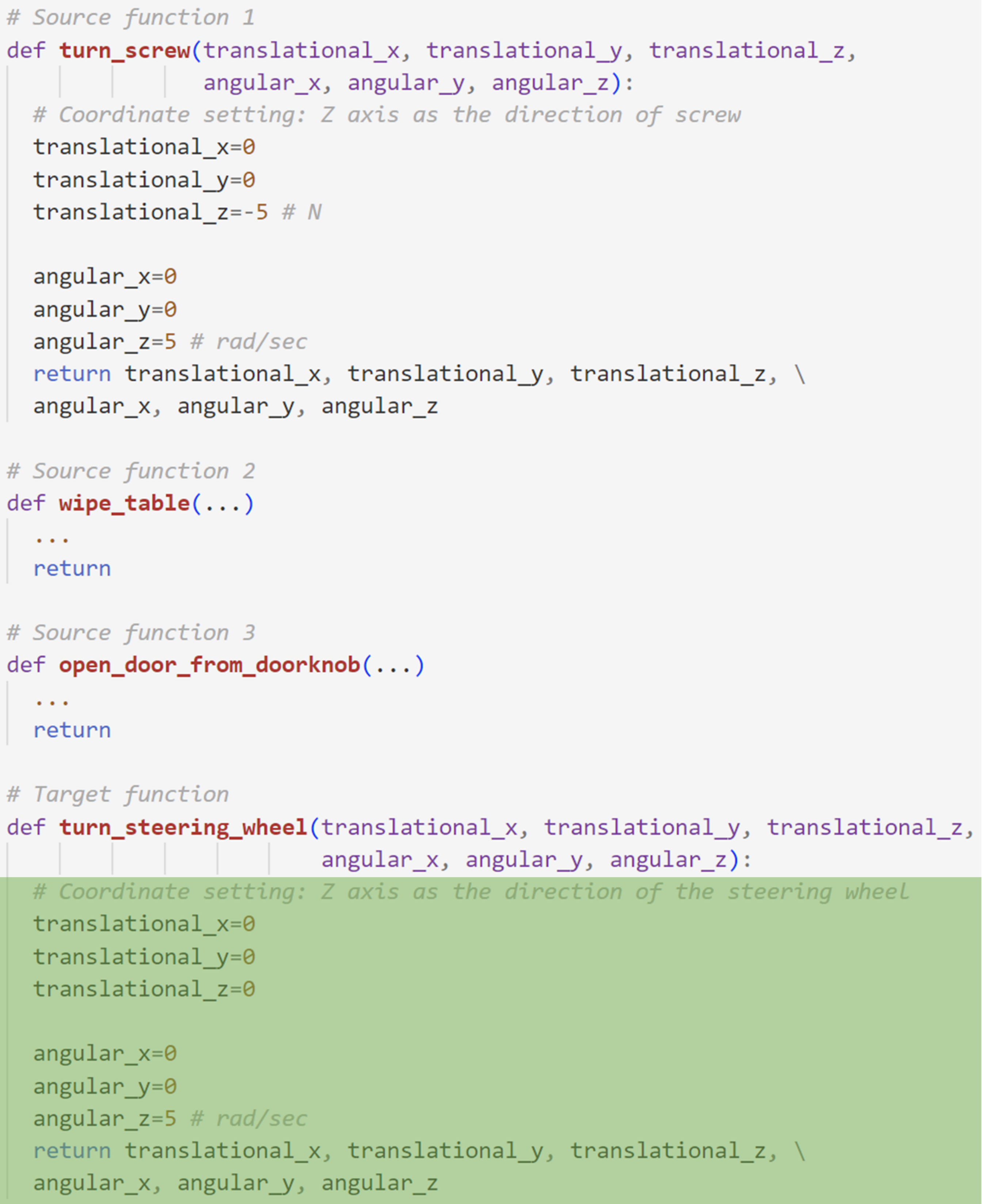}
     \caption{An exemplary 3-shot generation in Python syntax. The 3-shot prompt is shown in the grey background. Details in the source function 2 and 3 are skipped. The text generated by LLM is displayed in the green background. Note that the prompt can also be written in other common programming languages. More source functions are preferred. }
     \label{fig:TFFcode}
     \vspace{-3mm}
 \end{figure}

After obtaining a manipulator primitive task, we use LLMs to transit it to low-level execution in a TFF fashion. Since LLMs tend to produce lengthy answers and detailed explanations to casual questions, we need to design a specific prompt to regulate their output into a structured form. Inspired by the work~\cite{singh2022progprompt} that uses code blocks for robot-plan generation, we design our prompt in a function format in computer programming. The details of our prompt design are listed below, with a Python example in Figure~\ref{fig:TFFcode}.
\begin{itemize} 
\item Outside of the function, we use program comments to indicate whether a function is set for source specification or target generation. We recommend using multiple source functions with different TFF configurations to provide better guidance for the LLMs.

\item The function name is configured as the text of the primitive task. For example, the task ``turn screw'' is written as ``def turn\_screw(\enspace)''. Stop words such as ``a'' and ``the'' can be removed from the task name. In addition, the function parameters are specified as six-directional motions in the task frame formalism.
    
\item In the function, we first set the coordinates in the function comment. Then, we designate position/force set-points in a TFF manner. If one direction is inactive, we simply assign $0$ as its value. For the active directions, we use different units to distinguish whether they are  position- or force-controlled. For the example in Figure~\ref{fig:TFFcode}, we activate two out of the six directions using ``translational\_z=$-5$ \textit{\# N}'' and ``angular\_z=$5$ \textit{\# rad/sec}''. This specification implies that we rotate the screw about the $z$ axis at a velocity of 5 rad/sec while simultaneously applying a downward force of $5$ newtons. 

\item The target function sets the desired manipulator primitive task that we want to generate for. In this function, the target task name is the only element that needs to be filled. 
\end{itemize}

Once we input the whole prompt to a LLM, it will automatically generate the coordinate setting and specification of six-directional motions for the given target task. Using the generated TFF-based coordinates and set-points, robots can compute control commands based on their own manipulator configurations.

\vspace{-3mm}
\section{Evaluation}
In this section, we evaluate our TFF-based prompt with 4 state-of-the-art large language models -- OpenAI GPT-3.5-turbo, OpenAI GPT-4, Google Bard, and Meta LLaMA-2-70B.
\subsection{Evaluation Setting}
For the GPT-3.5-turbo and GPT-4 models, we applied the same setting. The temperature hyperparameter was set to be $0$ to minimize the randomness. The top P, frequency penalty, and presence penalty hyperparameters were configured to $1$, $0$, and $0$, respectively.

For the Google Bard model, there is no hyperparameter configuration available towards end-users. Thereby we cannot guarantee the deterministicness of the generated outcomes. We entered the prompt in its web GUI and only recorded the initial results generated.

The Meta  LLaMA-2 model is the most recently released LLM among them. We chose its 70B (70 billion) version, the one with the most parameters. We also set its temperature hyperparameter to the minimum value $0.01$. The repetition penalty was set to $1$ to grant repeated words with no penalty. Another hyperparameter top P was set to $1$.

We assessed the proposed prompt in a set of 30 manipulator primitive tasks as listed in Table~\ref{tab:table1}. We conducted tests in zero-shot, one-shot, three-shot and five-shot manners. The five-shot prompt was consisted of five source tasks as outlined in the block below. 

\begin{table*}[htb]
\centering
\begin{tabular}{l|l|l}
1. cut pizza & 11. rasp wood  & 21. open door from hinge\\
2. scrub desk with bench brush & 12. scrape substance from surface & 22. slide block over vertical surface\\
3. spear cake with fork & 13. peel potato & 23. turn steering wheel\\
4. fasten screw with screwdriver & 14. slice cucumber & 24. shake cocktail bottle\\
5. loosen screw with screwdriver & 15. flip bread & 25. cut banana\\
6. unlock lock with key & 16. shave object & 26. crack egg\\
7. fasten nut with wrench& 17. use roller to roll out dough & 27. press button\\
8. loosen nut with wrench& 18. insert peg into pegboard & 28. insert GPU into socket\\
9. spread paint with brush& 19. brush across tray & 29. open bottle\\
10. hammer in nail & 20. insert straw through cup lid & 30. open childproof bottle\\
\end{tabular}
\caption{Manipulator primitive tasks used in our evaluation. The tasks in the first two columns were selected from the Daily Interactive Manipulation dataset~\cite{PauliusE020} with minor modification, while the tasks listed in the third column were created by us.}
\label{tab:table1}
\end{table*}

\begin{mdframed}[backgroundcolor=gray!15]
\textit{\# Source function 1}\\
def turn\_screw(...)\\
  ...\\
\textit{\# Source function 2}\\
def wipe\_table(...)\\
  ...\\
\textit{\# Source function 3}\\
def open\_door\_from\_doorknob(...)\\
  ...\\
\textit{\# Source function 4}\\
def cut\_sandwich(...)\\
  ...\\
\textit{\# Source function 5}\\
def slide\_box\_upward\_on\_wall(...)\\
  ...
\end{mdframed}

For simplicity, we only show the task indicators and function names in this block while omitting other details. The three-shot prompt utilized the first three source tasks from the five-shot prompt, whereas the one-shot prompt only used the first one from it.

\subsection{Results}
The evaluations for OpenAI GPT and Google Bard were conducted in early July, 2023. The test for Meta LLAMA-2 was carried out in late July, 2023, shortly after its release. In order to validate the generated TFF specifications, We manually set up metrics for all 30 manipulator primitive tasks. The details of our metrics can be found in Appendix. If the TFF specification generated for a given task does not match the requirement specified in the metrics, we classify this generation as ``incorrect''. The overall correct rates are reported in Table~\ref{tab:table2}.
\begin{table}[htp]
\centering
\begin{tabular}{l|cccc}
{}   & 0-shot   & 1-shot     & 3-shot& 5-shot\\
\hline
GPT-3.5-turbo  & 0 & 0.30 & 0.70 & 0.67   \\
GPT-4 & 0  & 0.47   &  0.63 & 0.83\\
Bard  &  0 & 0 & 0.47    & 0.70  \\
LLaMA-2-70B & 0 & 0 & 0.07 & 0.13
\end{tabular}
\caption{Correct rates in the evaluation over 30 manipulator primitive tasks.}
 \label{tab:table2}
\end{table}

From the results, we notice that none of LLMs were able to produce a valid response in the zero-shot test. When we increased the number of source tasks to $5$, GPT-3.5-turbo, GPT-4, and Bard generated plenty of correct answers. Notably, the GPT-4 in the five-shot setting achieved the highest correct rate of $0.83$. Meanwhile, the LLaMA-2-70B failed the majority of tests. For the five-shot tests, the correctness of each manipulator primitive task is shown in Table~\ref{tab:table3}.

\begin{table}[htp]
    \centering
    \begin{tabular}{l|c|p{2cm}}
         &  Task No.& Correctness\\
         \hline
    \multirow{3}{*}{GPT-3.5-turbo}& 1-10&\tikz\draw[blue,fill=blue] (0,0) circle (.3ex); 
    \tikz\draw[blue,fill=blue] (0,0) circle (.3ex); 
    \tikz\draw[blue,fill=blue] (0,0) circle (.3ex); 
    \tikz\draw[blue,fill=blue] (0,0) circle (.3ex); 
    \tikz\draw[blue,fill=blue] (0,0) circle (.3ex); 
    \tikz\draw[red,fill=red] (0,0) circle (.3ex); 
    \tikz\draw[blue,fill=blue] (0,0) circle (.3ex); 
    \tikz\draw[blue,fill=blue] (0,0) circle (.3ex); 
    \tikz\draw[blue,fill=blue] (0,0) circle (.3ex); 
    \tikz\draw[blue,fill=blue] (0,0) circle (.3ex); 
    \\
     &11-20&\tikz\draw[red,fill=red] (0,0) circle (.3ex); 
    \tikz\draw[red,fill=red] (0,0) circle (.3ex); 
    \tikz\draw[red,fill=red] (0,0) circle (.3ex); 
    \tikz\draw[blue,fill=blue] (0,0) circle (.3ex); 
    \tikz\draw[blue,fill=blue] (0,0) circle (.3ex); 
    \tikz\draw[red,fill=red] (0,0) circle (.3ex); 
    \tikz\draw[red,fill=red] (0,0) circle (.3ex); 
    \tikz\draw[blue,fill=blue] (0,0) circle (.3ex); 
    \tikz\draw[blue,fill=blue] (0,0) circle (.3ex); 
    \tikz\draw[blue,fill=blue] (0,0) circle (.3ex); \\
     &21-30&\tikz\draw[blue,fill=blue] (0,0) circle (.3ex); 
    \tikz\draw[red,fill=red] (0,0) circle (.3ex); 
    \tikz\draw[blue,fill=blue] (0,0) circle (.3ex); 
    \tikz\draw[red,fill=red] (0,0) circle (.3ex); 
    \tikz\draw[blue,fill=blue] (0,0) circle (.3ex); 
    \tikz\draw[red,fill=red] (0,0) circle (.3ex); 
    \tikz\draw[blue,fill=blue] (0,0) circle (.3ex); 
    \tikz\draw[blue,fill=blue] (0,0) circle (.3ex); 
    \tikz\draw[blue,fill=blue] (0,0) circle (.3ex); 
    \tikz\draw[red,fill=red] (0,0) circle (.3ex); \\
     \hline 
    \multirow{3}{*}{GPT-4} & 1-10&\tikz\draw[blue,fill=blue] (0,0) circle (.3ex); 
    \tikz\draw[blue,fill=blue] (0,0) circle (.3ex); 
    \tikz\draw[blue,fill=blue] (0,0) circle (.3ex); 
    \tikz\draw[blue,fill=blue] (0,0) circle (.3ex); 
    \tikz\draw[blue,fill=blue] (0,0) circle (.3ex); 
    \tikz\draw[blue,fill=blue] (0,0) circle (.3ex); 
    \tikz\draw[blue,fill=blue] (0,0) circle (.3ex); 
    \tikz\draw[blue,fill=blue] (0,0) circle (.3ex); 
    \tikz\draw[blue,fill=blue] (0,0) circle (.3ex); 
    \tikz\draw[blue,fill=blue] (0,0) circle (.3ex); 
    \\
    & 11-20&\tikz\draw[blue,fill=blue] (0,0) circle (.3ex); 
    \tikz\draw[blue,fill=blue] (0,0) circle (.3ex); 
    \tikz\draw[red,fill=red] (0,0) circle (.3ex); 
    \tikz\draw[blue,fill=blue] (0,0) circle (.3ex); 
    \tikz\draw[blue,fill=blue] (0,0) circle (.3ex); 
    \tikz\draw[red,fill=red] (0,0) circle (.3ex); 
    \tikz\draw[blue,fill=blue] (0,0) circle (.3ex); 
    \tikz\draw[blue,fill=blue] (0,0) circle (.3ex); 
    \tikz\draw[blue,fill=blue] (0,0) circle (.3ex); 
    \tikz\draw[red,fill=red] (0,0) circle (.3ex); 
    \\
    & 21-30&\tikz\draw[blue,fill=blue] (0,0) circle (.3ex); 
    \tikz\draw[blue,fill=blue] (0,0) circle (.3ex); 
    \tikz\draw[blue,fill=blue] (0,0) circle (.3ex); 
    \tikz\draw[blue,fill=blue] (0,0) circle (.3ex); 
    \tikz\draw[blue,fill=blue] (0,0) circle (.3ex); 
    \tikz\draw[red,fill=red] (0,0) circle (.3ex); 
    \tikz\draw[blue,fill=blue] (0,0) circle (.3ex); 
    \tikz\draw[blue,fill=blue] (0,0) circle (.3ex); 
    \tikz\draw[blue,fill=blue] (0,0) circle (.3ex); 
    \tikz\draw[red,fill=red] (0,0) circle (.3ex); 
    \\
    \hline
    \multirow{3}{*}{Bard}   & 1-10&\tikz\draw[blue,fill=blue] (0,0) circle (.3ex); 
    \tikz\draw[red,fill=red] (0,0) circle (.3ex); 
    \tikz\draw[blue,fill=blue] (0,0) circle (.3ex); 
    \tikz\draw[blue,fill=blue] (0,0) circle (.3ex); 
    \tikz\draw[blue,fill=blue] (0,0) circle (.3ex); 
    \tikz\draw[blue,fill=blue] (0,0) circle (.3ex); 
    \tikz\draw[blue,fill=blue] (0,0) circle (.3ex); 
    \tikz\draw[red,fill=red] (0,0) circle (.3ex); 
    \tikz\draw[blue,fill=blue] (0,0) circle (.3ex); 
    \tikz\draw[blue,fill=blue] (0,0) circle (.3ex); \\
    & 11-20&\tikz\draw[red,fill=red] (0,0) circle (.3ex); 
    \tikz\draw[red,fill=red] (0,0) circle (.3ex); 
    \tikz\draw[blue,fill=blue] (0,0) circle (.3ex); 
    \tikz\draw[blue,fill=blue] (0,0) circle (.3ex); 
    \tikz\draw[blue,fill=blue] (0,0) circle (.3ex); 
    \tikz\draw[blue,fill=blue] (0,0) circle (.3ex); 
    \tikz\draw[red,fill=red] (0,0) circle (.3ex); 
    \tikz\draw[blue,fill=blue] (0,0) circle (.3ex); 
    \tikz\draw[blue,fill=blue] (0,0) circle (.3ex); 
    \tikz\draw[red,fill=red] (0,0) circle (.3ex); \\
    & 21-30&\tikz\draw[blue,fill=blue] (0,0) circle (.3ex); 
    \tikz\draw[red,fill=red] (0,0) circle (.3ex); 
    \tikz\draw[blue,fill=blue] (0,0) circle (.3ex); 
    \tikz\draw[blue,fill=blue] (0,0) circle (.3ex); 
    \tikz\draw[red,fill=red] (0,0) circle (.3ex); 
    \tikz\draw[red,fill=red] (0,0) circle (.3ex); 
    \tikz\draw[blue,fill=blue] (0,0) circle (.3ex); 
    \tikz\draw[blue,fill=blue] (0,0) circle (.3ex); 
    \tikz\draw[blue,fill=blue] (0,0) circle (.3ex); 
    \tikz\draw[blue,fill=blue] (0,0) circle (.3ex); 
    \\ \hline
    \multirow{3}{*}{LLaMA-2-70B}   
    &1-10&\tikz\draw[blue,fill=blue] (0,0) circle (.3ex); 
    \tikz\draw[red,fill=red] (0,0) circle (.3ex); 
    \tikz\draw[red,fill=red] (0,0) circle (.3ex); 
    \tikz\draw[red,fill=red] (0,0) circle (.3ex); 
    \tikz\draw[blue,fill=blue] (0,0) circle (.3ex); 
    \tikz\draw[red,fill=red] (0,0) circle (.3ex); 
    \tikz\draw[red,fill=red] (0,0) circle (.3ex); 
    \tikz\draw[red,fill=red] (0,0) circle (.3ex); 
    \tikz\draw[red,fill=red] (0,0) circle (.3ex); 
    \tikz\draw[red,fill=red] (0,0) circle (.3ex); \\
    &11-20&\tikz\draw[red,fill=red] (0,0) circle (.3ex); 
    \tikz\draw[red,fill=red] (0,0) circle (.3ex); 
    \tikz\draw[red,fill=red] (0,0) circle (.3ex); 
    \tikz\draw[red,fill=red] (0,0) circle (.3ex); 
    \tikz\draw[blue,fill=blue] (0,0) circle (.3ex); 
    \tikz\draw[red,fill=red] (0,0) circle (.3ex); 
    \tikz\draw[red,fill=red] (0,0) circle (.3ex); 
    \tikz\draw[red,fill=red] (0,0) circle (.3ex); 
    \tikz\draw[red,fill=red] (0,0) circle (.3ex); 
    \tikz\draw[red,fill=red] (0,0) circle (.3ex); \\
    &21-30&\tikz\draw[red,fill=red] (0,0) circle (.3ex); 
    \tikz\draw[red,fill=red] (0,0) circle (.3ex); 
    \tikz\draw[red,fill=red] (0,0) circle (.3ex); 
    \tikz\draw[red,fill=red] (0,0) circle (.3ex); 
    \tikz\draw[blue,fill=blue] (0,0) circle (.3ex); 
    \tikz\draw[red,fill=red] (0,0) circle (.3ex); 
    \tikz\draw[red,fill=red] (0,0) circle (.3ex); 
    \tikz\draw[red,fill=red] (0,0) circle (.3ex); 
    \tikz\draw[red,fill=red] (0,0) circle (.3ex); 
    \tikz\draw[red,fill=red] (0,0) circle (.3ex); \\ \hline
    \end{tabular}
    \caption{The correctness of each manipulator primitive task in the 5-shot tests. Blue dots indicate the correct generation for tasks, while red dots represent incorrect ones. Task No. corresponds to the numbering in Table~\ref{tab:table1}, with every 10 tasks sequentially lined-up in a row.}
     \label{tab:table3}
\end{table}
For GPT-3.5-turbo, GPT-4, and Bard models, they strictly followed the desired ``\textit{\# Coordinate setting}; translational ...; angular ... '' format even since the one-shot prompt. Both GPT models can also figure out some task specification just depending on the one-shot prompt. For the Bard in the one-shot setting, most of its target task specifications essentially duplicated the direction setting from the source task ``turn screw.'' The only improvisational part was the new object regarding the $z$ axis in the coordinate setting. 

Next, we take a close look into the GPT-4 five-shot case, the one with the best performance. In total, its five failed cases were ``peel potato'', ``shave object'', ``insert straw through cup lid'', ``crack egg'', and ``open childproof bottle.'' These failed tasks are quite divergent from the source tasks and pose more challenges to LLMs. Take the ``insert straw through cup lid'' task as an example, the GPT-4 generated a single active direction ``translational\_z = -5  \textit{\# N}'', which implied using force control in the straw-insertion direction. However, this task should be position controlled because the straw has no interaction with any outside rigid object and there is only minor friction between the straw and the cup lid. For another failed task ``open childproof bottle'', it is supposed to have the opposite force direction compared to the normal ``open bottle'' task. However, the GPT-4 model generated an identical answer for these two tasks.

Regarding the LLaMA-2-70B, our current prompt has difficulty in guiding it to the desired specification. Even up to the 5-shot prompt, it barely generated any suitable response. One typical response is shown below. Note that some redundant details of the function parameters are skipped using ``...''. 
\begin{mdframed}[backgroundcolor=gray!15]
  \textit{\# Combine the five source functions}\\
  turn\_screw(...)\\
  wipe\_table(...)\\
  open\_door\_from\_doorknob(...)\\
  cut\_sandwich(...)\\
  slide\_box\_upward\_on\_wall(...)
\end{mdframed}
The LLaMA-2-70B just repeatedly called the source functions. It was not capable of generating any new coordinate setting or position/force set-points with our prompt. 

In addition to focusing our attention on the direction settings, we also inspected the numerical values that were generated. We used values such as $5$ (rad/sec), $5$ (N) and $2$ (cm) in the prompt. Most of the generated values were close to these ones and sometimes reached $10$. Among all generated values in the 5-shot tests, we noticed two unusual values, both produced by Bard -- 1) ``hammer in nail'' task: ``translational\_z = 50  \textit{\# N}'', 2) ``cut banana'' task: ``angular\_z = 100 \textit{\#  rad/sec}''. In the hammer case, it is reasonable to set a large force. However, the direction setting the banana cutting case is incorrect, and more notably, the generated $100$ (rad/s) set-point is excessively high for common manipulators to execute. This raises an additional concern: if the LLMs generate abnormal set-point values, might it potentially result in safety issues in robot execution?

\section{Conclusions and Discussions}
This paper presented a LLM-based approach to convert textual manipulator primitive tasks to TFF-based position/force set-points. By integrating the generalizability of LLMs and the cross-task transferability of TFF together, we have provided a new solution for the planning-to-execution transition problem in layered architectures. The evaluations showed the effectiveness of our prompt in three LLMs -- OpenAI GPT-3.5-turbo, OpenAI GPT-4, and Google Bard.

Since the correct rate increases as more source tasks are offered, we suggest building a more comprehensive prompt that includes tasks in all contact and movement types. In addition, further study can be conducted to guide the LLaMA-2 model in producing correct specifications. 

Furthermore, the evaluation on the LLM-generated content remains a challenging problem. We believe that there exists a necessity to design a comprehensive and standardized evaluation set for assessing the language grounding to actions. The recent work of Google RT-1~\cite{DBLP:conf/rss/BrohanBCCDFGHHH23} established a large-scale real-world dataset connecting images and natural language instructions with manipulator actions but is limited to a few verbs. In the future, we will establish an evaluation set that includes a broader range of verbs that can better capture the contact-rich characteristics of manipulators.

\appendix
\section{Appendix: Evaluation Metrics }
In this appendix, we provide detailed metrics used in the Evaluation section. In certain cases, the $x$, $y$, and $z$ axes can be interchangeable depending on the coordinate setting.
\begin{table}[htb]
\centering
\begin{tabularx}{\linewidth}{|X|}
\multicolumn{1}{c}{\textbf{TFF Requirements for Primitive Tasks}}\\ \hline
1. cut pizza: 1 translational direction activated. \\  
2. scrub desk with bench brush: 1 or 2 translational directions activated, must apply force on plane. \\
3. spear cake with fork: only $z$ translational direction activated. \\
4. fasten screw with screwdriver: $z$ angular direction activated. \\ 
5. loosen screw with screwdriver: opposite as Task 4.\\
6. unlock lock with key: $z$ angular direction activated. \\
7. fasten nut with wrench: $z$ angular direction activated.\\
8. loosen nut with wrench: opposite as Task 7.\\
9. spread paint with brush: translational direction on $x-y$ plane  activated.  \\
10. hammer in nail: only $z$ translational direction activated.\\
11. rasp wood: same as Task 2. \\
12. scrape substance from surface: same as Task 2.\\
13. peel potato: 1 translational direction activated, must apply force on plane.\\
14. slice cucumber: same as Task 1.\\
15. flip bread: 1 angular direction activated.\\
16. shave object: same as Task 2. \\
17. use roller to roll out dough: same as Task 2.\\
18. insert peg into pegboard: same as Task 3.\\
19. brush across tray: same as Task 2 or Task 9.\\
20. insert straw through cup lid: 1 translational direction activated, position control.\\
21. open door from hinge: only $z$ angular direction activated.\\
22. slide block over vertical surface: $z$ translational direction activated, must apply force on $x-y$ plane. \\
23. turn steering wheel: same as Task 6.\\
24. shake cocktail bottle: angular direction activated, position control.\\
25. cut banana: same as Task 1.\\
26. crack egg: same as Task 20.\\
27. press button: same as Task 3.\\
28. insert GPU into socket: same as Task 3.\\
29. open bottle: same as Task 4.\\
30. open childproof bottle: add downside force versus Task 29.\\
\hline
\end{tabularx}
\end{table}

\bibliography{aaai23}

\end{document}